\documentclass[conference]{IEEEtran}
\IEEEoverridecommandlockouts
\usepackage{cite}
\usepackage{amsmath,amssymb,amsfonts}
\usepackage{algorithmic}
\usepackage{graphicx}
\usepackage{textcomp}
\usepackage{xcolor}
\usepackage{multirow}
\usepackage[table]{xcolor}
\usepackage{algorithm}
\usepackage{algorithmic}
\usepackage[dvipsnames]{xcolor}
\usepackage{subcaption}
\usepackage[shortlabels]{enumitem}
\usepackage[capitalize]{cleveref}
\usepackage{xspace}
\usepackage{balance}
\usepackage{hhline}

\newcommand{\clientidx}{\ensuremath{k}}
\newcommand{\algname}{\texttt{\textsc{ProDiGy}}\xspace}
\newcommand{\model}{\ensuremath{\boldsymbol{\theta}}}
\DeclareMathOperator{\argmin}{\mathrm{argmin}}

\def\BibTeX{{\rm B\kern-.05em{\sc i\kern-.025em b}\kern-.08em
    T\kern-.1667em\lower.7ex\hbox{E}\kern-.125emX}}
\begin{document}

\title{\algname: Proximity- and Dissimilarity-Based Byzantine-Robust Federated Learning}

\author{\IEEEauthorblockN{Sena Ergisi, Luis Maßny, Rawad Bitar}
\IEEEauthorblockA{
School of Computation, Information and Technology,
Technical University of Munich, Germany\\
\{sena.ergisi, luis.massny, rawad.bitar\}@tum.de} \thanks{This project is funded by the German Research Foundation (DFG) under Grant Agreement Nos. BI 2492/1-1 and KR 3517/13-1. Sena Ergisi is partly supported by the joint scholarship program of the German Academic Exchange Service (DAAD) and the Turkish Education Foundation (TEV).}
\vspace{-1cm}}

\maketitle

\begin{abstract}
Federated Learning (FL) emerged as a widely studied paradigm for distributed learning. Despite its many advantages, FL remains vulnerable to adversarial attacks, especially under data heterogeneity. We propose a new Byzantine-robust FL algorithm called \algname. The key novelty lies in evaluating the client gradients using a joint dual scoring system based on the gradients' \texttt{Pro}ximity and \texttt{Di}ssimilarity. We demonstrate through extensive numerical experiments that \algname outperforms existing defenses in various scenarios. In particular, when the clients' data do not follow an IID distribution, while other defense mechanisms fail, \algname maintains strong defense capabilities and model accuracy. These findings highlight the effectiveness of a dual perspective approach that promotes natural similarity among honest clients while detecting suspicious uniformity as a potential indicator of an attack. 
\end{abstract}

\begin{IEEEkeywords}
Federated Learning, Byzantine Robustness, Data Heterogeneity, Adversarial Attacks, Robust Aggregation
\end{IEEEkeywords}

\section{Introduction}

Federated Learning (FL)~\cite{mcmahan} has many advantages over centralized learning methods. The necessity of collecting data on a single machine disappears, and processing private data without direct access becomes applicable. Despite the many advantages, FL also faces several challenges, including preserving the privacy of the clients' data, efficiency of communication over resource-limited networks, and security (robustness) against adversarial clients~\cite{openprobs,survey1}. In particular, those challenges are exacerbated in the practical setting where the clients' data are heterogeneous, i.e., the clients' data are not necessarily drawn from the same distribution, see, e.g.~\cite{heterogeneousfl}.

In a typical FL communication round, a central server broadcasts the current global model parameters to participating clients, which subsequently return local updates. The server then aggregates these updates to produce the updated global model. The distributed architecture of FL makes it particularly vulnerable to Byzantine\footnote{Malicious clients are called Byzantine in reference to the Byzantine generals problem~\cite{byz}.} clients, i.e., clients that intentionally jam the learning process. It is shown in~\cite{krum} that, if care is not taken, only one Byzantine client can hinder the convergence of the training process. 

Byzantine-robust aggregation methods are shown to effectively mitigate the effect of Byzantine clients \cite{mediantrimmean,geomed,rfa,centeredclipping,krum, bulyan, faba, sniper, clustering1, clustering2, clippedclustering, zeno, fltrust, fedgt}. Their robustness guarantees are further enhanced by incorporating various techniques, such as pre-aggregation~\cite{fixbymix,mimic} and local momentum~\cite{momentummhamdi}. While these defenses perform effectively in homogeneous settings, their robustness significantly degrades under heterogeneous conditions. In such scenarios, model convergence failures under state-of-the-art attacks remain a pressing concern. The main reason is that data heterogeneity among clients introduces a gradient similarity gap that can be exploited by adversaries, rendering Byzantine updates increasingly indistinguishable from honest updates. This challenge is further exacerbated in the presence of colluding adversaries, who can craft harmful updates that bypass the robust aggregation and bias the global gradient into a desired direction, thereby amplifying their detrimental impact on the global model~\cite{mimic, alie, foe}.

To mitigate the effect of Byzantine clients even under high data heterogeneity, we propose a novel robust aggregation method, called \algname. The method assesses the similarity of the clients' gradients from two complementary perspectives. First, it leverages the proximity of honest updates to assign reliability scores to clients, an approach consistent with many established defenses. Second, and more critically, it constrains gradient similarity to prevent adversarial updates from aligning in a single direction, thereby mitigating a potential bias in the gradient direction.

We evaluate \algname across diverse experimental settings\footnote{The code is provided at github.com/senaergisi/prodigy.} and compare it against prominent Byzantine-robust aggregation schemes. We show a significant improvement in mitigating the effect of Byzantine clients. In settings where state-of-the-art robust aggregation rules fail, \algname still maintains model utility, cf. \cref{tab:bigtable,tab:femnistt,tab:femnistt100}. In all other settings, \algname provides higher worst-case utility than prominent robust aggregation rules, cf. \cref{tab:table1,tab:table2}.

\section{Related Work}

\subsection{Robust Federated Learning}
It has been shown that linear aggregation rules are susceptible to Byzantine attacks~\cite{krum}. A popular countermeasure is to use robust aggregation rules that analyze the properties, e.g., statistics and distances, of the clients' gradients. Robust aggregation methods have various approaches to mitigate the effect of adversarial attacks~\cite{challengesandapproaches}, but they often rely on the common assumption that honest model updates are similar, whereas malicious updates appear as outliers. \emph{Statistics-based defenses} use the statistical properties of gradients to obtain the best representative of the honest gradients while excluding outliers~\cite{mediantrimmean,geomed,rfa,centeredclipping}. \emph{Distance-based defenses} leverage similarities among gradients to distinguish honest gradients from malicious ones. Pairwise Euclidean distances~\cite{krum, bulyan, faba, sniper} and cosine similarity~\cite{clustering1, clustering2, clippedclustering} are the most popular metrics used to reveal the similarity among gradients. \emph{Performance-based defenses} assess the quality of client updates, typically requiring access to a clean validation dataset at the server side~\cite{zeno, fltrust,fedgt}. Although we do not assume that the server has access to a clean dataset, \algname shares conceptual similarities with approaches that assign trust scores to client gradients.

Beyond Byzantine threats, Sybil attacks~\cite{sybil} pose a significant challenge by allowing adversaries to arbitrarily join or leave the system, often simulating multiple clients to increase their disruptive influence. Unlike Byzantine defenses that treat gradient similarity as benign, Sybil defenses assume that such similarity may indicate malicious collusion, see, e.g., \cite{foolsgold}. 

\subsection{Heterogeneous Federated Learning}

Heterogeneity presents a fundamental challenge in FL, encompassing differences in data distributions, model structures, and communication networks \cite{heterogeneousfl}. Statistical-level and system-level imbalance have been explored, e.g.,\cite{fedprox, scaffold}. Data heterogeneity particularly causes divergence among honest gradients, as local and global objectives no longer align. This is because, in a heterogeneous setting, data at different clients follow distributions that are not necessarily the same, commonly referred to as a non-IID data distribution. Severe data heterogeneity can resemble adversarial behavior, leading to significant degradation in global model performance even in the absence of malicious clients~\cite{mimic}.

Byzantine-robust aggregation methods perform best when honest gradients exhibit low divergence. \emph{Pre-aggregation} techniques such as \emph{Bucketing}~\cite{mimic} and \emph{Nearest Neighbor Mixing (NNM)}~\cite{fixbymix} aim to reduce gradient divergence through random or structured mixing, thereby enhancing the effectiveness of subsequent aggregation. While NNM is computationally more expensive than Bucketing, it offers greater robustness improvements. History of the client gradients offers additional insights into client behavior and is leveraged by several robust aggregation methods, e.g., \cite{foolsgold}, \cite{centeredclipping}. An alternative approach to incorporating history is through \emph{local momentum}~\cite{momentummhamdi}, which enhances learning robustness also under client heterogeneity.

\section{System Model and Problem Description}

\subsection{Learning Objective} 

We consider an FL system with a central server and $N$ clients. The clients possess datasets over which the server wants to compute a function, e.g., train a neural network. Let $\mathcal{S}^{(1)}, \dots, \mathcal{S}^{(N)} \subseteq \mathcal{X}\times\mathcal{Y}$ denote the $N$ clients' local datasets, $\mathcal{X}$ denotes the input space of the data and $\mathcal{Y}$ denotes the output (label) space. We assume the datasets are drawn from (potentially non-IID) distributions $\mathcal{D}^{(1)},\dots,\mathcal{D}^{(N)}$. We denote by $\mathcal{S} \triangleq \cup_{k\in[N]} \mathcal{S}^{(k)}$ the global dataset, where we define $[N] \triangleq \{1,\dots,N\}$.

The learning objective is to find a function $\mathcal{F}(\boldsymbol{\theta};\mathbf{x})$ parametrized by a vector $\boldsymbol{\theta}$ that correctly estimates the label of samples from $\mathcal{S}$, i.e., for any $(\mathbf{x}, y)\in \mathcal{X} \times \mathcal{Y}$, the function $\mathcal{F}(\boldsymbol{\theta};\mathbf{x})$ outputs $\hat{y}$ and the goal is for $\hat{y}$ to be equal to $y$. To measure how far the output of $\mathcal{F}(\boldsymbol{\theta};\mathbf{x})$ is from $y$, for $k\in [N]$, a per-client loss function is defined as
\begin{equation*}
    \mathcal{L}_k(\boldsymbol{\theta}) = \frac{1}{|\mathcal{S}^{(k)}|}\sum_{(\boldsymbol{x}, y) \in \mathcal{S}^{(k)}} \ell(\mathcal{F}(\boldsymbol{\theta};\mathbf{x}), y),
\end{equation*}
where $\ell: \mathbb{R}^d \times \mathcal{X}\times \mathcal{Y} \rightarrow \mathbb{R}$ is a point-wise loss function assumed to be differentiable with respect to $\boldsymbol{\theta}$. The goal of the central server is to find the best parameter vector $\boldsymbol{\theta}^\star$ that minimizes the loss function defined as
\begin{equation*}
    \mathcal{L}(\boldsymbol{\theta})\triangleq \frac{1}{|\mathcal{S}|}\sum_{k=1}^{N} |\mathcal{S}^{(k)}|\mathcal{L}_k(\boldsymbol{\theta}).
\label{eq:ermfl}
\end{equation*}

In other words, the central server wants to solve the following optimization problem
   $$ \boldsymbol{\theta}^\star = \argmin_{\boldsymbol{\theta}\in \mathbb{R}^d} \mathcal{L}(\boldsymbol{\theta}).$$

To find $\model^\star$, the server employs the iterative mini-batch gradient descent method starting from a random vector $\model_0$. At each round $t$, client $k$ computes and sends a model update $\boldsymbol{g}^k_t$ based on a subset of its local data and the gradient of the loss function with respect to $\model$ (see lines 10--15 in \cref{fl_alg}). Some variants of the stochastic gradient descent apply local momentum with parameter $\beta\in [0,1]$. In such variants, instead of sending $\boldsymbol{g}^k_t$ to the server, client $k$ compute and sends a local momentum $\boldsymbol{m}^k_t$ recursively updated as
\begin{equation*}
    \boldsymbol{m}^k_t = \beta \boldsymbol{m}^k_{t-1} + (1-\beta)\boldsymbol{g}^k_t.
\end{equation*}
The server aggregates the obtained local momenta according to a certain aggregation rule $\texttt{Agg}(\cdot)$ (typically weighted average) to update the global model as $\model_{t+1} = \model_t - \gamma_t \texttt{Agg}(\boldsymbol{m}^1_t,\dots,\boldsymbol{m}^N_t)$. The process repeats for $T$ rounds or until certain convergence criteria are met. 

\begin{algorithm}[!t]
\caption{Federated Learning Algorithm}
\label{fl_alg}
\begin{algorithmic}[1]
\STATE \textbf{Input:} total number of clients $N$, initial model parameters $\boldsymbol{\theta}_0 \in \mathbb{R}^d$, total communication rounds  $T$, learning rate in round $t$  $\gamma_t$, local iteration number $E$, batch size $b$\\
\vspace{5pt}
\textbf{Server:}
\vspace{5pt}
\FOR{\text{round} $t \in \{0,\dots,T-1\} $ } 
    \STATE Broadcast $\boldsymbol{\theta}_t$ to clients
    \FOR{each honest client $k\in[N]$ in parallel} 

    \STATE $\boldsymbol{g}^k_t \gets \text{ClientUpdate}(k, \boldsymbol{\theta}_t, t)$ 
    \ENDFOR
    \STATE Aggregate $\boldsymbol{\theta}_{t+1} \gets \boldsymbol{\theta}_{t} - {\gamma_t} \texttt{Agg}(\boldsymbol{g}^1_{t},\dots, \boldsymbol{g}^N_{t})$ \label{line:agg}
\ENDFOR \\
\vspace{5pt}
\textbf{ClientUpdate $(k, \boldsymbol{\theta}_t,t)$:}\\
\vspace{5pt}
\STATE Initialize $\boldsymbol{\theta}^{(0)} = \boldsymbol{\theta}_t$
\FOR{$e \in \{0,\dots,E-1\}$}
    \STATE Sample randomly and independently without replacement a set of data points $\mathcal{S}^{(k)}_{e}$ of size $b$ from the local dataset $\mathcal{S}^{(k)}$ 
    \STATE $\nabla\boldsymbol{\theta}^{(e)} \gets \frac{1}{b}\sum_{(\mathbf{x}, y) \in \mathcal{S}^{(k)}_{e}} \nabla \ell(\mathcal{F}(\boldsymbol{\theta}^{(e)};\mathbf{x}), y)$
    \STATE $\boldsymbol{\theta}^{(e+1)} \gets \boldsymbol{\theta}^{(e)}  - \gamma_t\nabla\boldsymbol{\theta}^{(e)}$ \label{line:localgradientupdate}
\ENDFOR
\STATE Send $\boldsymbol{g}^k_t \gets \frac{\boldsymbol{\theta}_t-\boldsymbol{\theta}^{(E)}}{\gamma_t}$ to the server
\vspace{5pt}
\STATE \textbf{return} $\boldsymbol{\theta}_T$
\end{algorithmic}
\end{algorithm}

\subsection{Adversarial Model}

We assume the presence of $f<\frac{N}{2}$ Byzantine clients maliciously trying to jam the learning process. We use the strongest adversarial model, where each client has full knowledge of the learning process, including the aggregation method that the server uses and the updates of the honest clients. In addition, adversarial clients can collude to design a joint attack. In such a setting, the optimization task changes to 
\begin{equation*}
    \min_{\boldsymbol{\theta}\in \mathbb{R}^d} \frac{1}{|\mathcal{S}^{(\mathcal{H})}|}\sum_{k\in \mathcal{H}} |\mathcal{S}^{(k)}|\mathcal{L}_k(\boldsymbol{\theta}),
\label{eq:ermbyzfl}
\end{equation*}
where $\mathcal{H}$ denotes the set of honest clients and $\mathcal{S}^{(\mathcal{H})} \triangleq \cup_{k\in \mathcal{H}} \mathcal{S}^{(k)} $.

\subsection{Robust Aggregation}
To mitigate the effect of the Byzantine clients, the server needs to carefully craft a \emph{robust aggregation rule} that takes as input the $N$ updates, out of which $f$ might be corrupt, and outputs an aggregated update that is faithful to the $N-f$ honest updates. Examples of such rules are computing the mean or the median of the input vectors, assuming that the majority of the vectors are honest, the output will be faithful to the honest updates. Several robust aggregation rules are given in \cref{sec:setup}.

Here, we assume the server does not have a clean data set on which it can test the performance of the updates sent by clients. It also has no control over the training data of the clients and no information on the distribution of the data. The only information it can leverage is $N$ and $f$. Upon receiving $N$ local updates $(\boldsymbol{g}^1_{t},\dots, \boldsymbol{g}^N_{t})$, the server applies an aggregation rule $\texttt{Agg}(\cdot)$, see Line \ref{line:agg} in Algorithm \ref{fl_alg}, that is designed to be robust. The overall FL algorithm used in this paper is given in \cref{fl_alg}.

The crux of a Byzantine-robust FL algorithm is designing a robust aggregation rule that can be computed based on the information available to the server and allows for the best accuracy given any adversarial attack.

\section{\algname}
\label{sec:prodigy}

We propose a robust aggregation rule with \texttt{Pro}ximity and \texttt{Di}ssimilarity based \texttt{G}radient scoring (\algname). Similar to previous methods, \algname computes a trust score $s(k)$ for each client $k$ and outputs an average of local updates weighted by the trust scores:
\begin{equation*}
     \algname(\boldsymbol{g}^1_t, \dots, \boldsymbol{g}^N_t) = \frac{\sum_{k \in [N]} s(k)\cdot\boldsymbol{g}^k_t}{\sum_{k \in [N]}s(k)}.
\end{equation*}
This function replaces the aggregation step $\texttt{Agg}(\boldsymbol{g}^1_{t},\dots, \boldsymbol{g}^N_{t})$ in \cref{fl_alg} in each round. For ease of presentation, we omit the round index $t$ when it is clear from the context.

The main innovation is a composite trust score $s(\clientidx)$, which consists of two complementary components $s_p(k)$ and $s_d(k)$. On a high level, the \emph{proximity} score $s_p(\cdot)$ favors the proximity between local updates, using pairwise squared Euclidean distances. 
The \emph{dissimilarity score} $s_d(\cdot)$ favors the dissimilarity in the neighborhood of an honest local update.
For each client, \algname computes the two scores $s_p(k)$ and $s_d(k)$ and a threshold value $s_{th}$, and combines the scores according to
\begin{equation}
\label{eq:final_score}
s(k) =
\begin{cases}
      0, & \text{ if } s_p(k) \cdot s_d(k) \leq s_{th},\\
     s_p(k) \cdot s_d(k), & \text{ otherwise}. 
\end{cases}
\end{equation}
The overall procedure is provided in \cref{prodigy_algorithm}.

The server computes the pairwise $\ell_2$-distance of all client updates, i.e., $\left\|\boldsymbol{g}^k-\boldsymbol{g}^{k^\prime} \right\|^2 $ for all $k\neq k^\prime \in [N]$. For each client $k$, the server sorts and indexes the other clients according to their update's distance. In other words, we denote by $k_i$ the $i$-th closest neighbor of client $k$ and therefore we have $\delta_{k,1}\leq \delta_{k,2} \leq\dots\leq \delta_{k, {N-1}}$, where $\delta_{k,i} \triangleq \left\|\boldsymbol{g}^k-\boldsymbol{g}^{k_i} \right\|^2 $. In addition, for each client $k$, we define the neighborhood of closest $f$ clients including the client $k$ itself as $\mathcal{N}_{f}(k) \triangleq \{k , {k_1},\dots, {k_{f-1}}\}$.

\algname computes the proximity score for client $k$ as:
\begin{equation}\label{eq:score1}
    s_p(k) = \frac{1}{\sum_{i = f}^{N-f-1} \delta_{k,i}}.
\end{equation}

Subsequently, the dissimilarity score for client $k$ is
\begin{equation}\label{eq:score2}
    s_d(k) = \frac{{\sigma}(\mathcal{N}_{f}(k))}{\|\boldsymbol{\mu}(\mathcal{N}_{f}(k))\|}, 
\end{equation}
where $\boldsymbol{\mu}(\mathcal{N}_{f}(k)) = \frac{1}{{|\mathcal{N}_{f}(k)|}}\sum_{i \in \mathcal{N}_{f}(k)} \boldsymbol{g}^i$ denotes the mean of the updates in the set $\mathcal{N}_{f}(k)$ and
${\sigma}(\mathcal{N}_{f}(k)) = \sqrt{\frac{1}{|\mathcal{N}_{f}(k)|} \sum_{i \in \mathcal{N}_{f}(k)} \left\|\boldsymbol{g}^i - \boldsymbol{\mu}(\mathcal{N}_{f}(k))\right\|^2}$ denotes their standard deviation. The dissimilarity score is equivalent to the \emph{coefficient of variance} of the gradients in $\mathcal{N}_{f}(k)$.

After having computed the proximity and dissimilarity score, \algname eliminates the $f$ clients that have the lowest scores. Formally, denoting by $j_i$ the client with the $i$-th smallest composite score $s^\prime(j_i)=s_p(j_i) \cdot s_d(j_i)$, \algname computes a threshold value $s_{th}=s(j_f)$ and applies \cref{eq:final_score} to filter out the clients with lowest scores $s^\prime(k)$.

\begin{algorithm}[!t]
\caption{\algname Algorithm}
\label{prodigy_algorithm}
\begin{algorithmic}[1]
\STATE \textbf{Input:} Gradient estimates sent by clients $\boldsymbol{g}^1, \dots, \boldsymbol{g}^N \in \mathbb{R}^d$, total number of clients $N$, number of Byzantine clients $f$

\FOR{each client $k\in [N]$ } 
    \STATE $s_p(k)\gets \frac{1}{\sum_{i = f}^{N-f-1} \delta_{k,i}}$
    \vspace{3pt}
    \STATE $s_d(k) \gets \frac{{\sigma}(\mathcal{N}_{f}(k))}{\|\boldsymbol{\mu}(\mathcal{N}_{f}(k))\|}$
\ENDFOR
\STATE $s_{th} \gets s_p(j_f) \cdot s_d(j_f)$
\STATE 
$s(k) \gets
\begin{cases}
      0, & \text{ if } s_p(k) \cdot s_d(k) \leq s_{th},\\
     s_p(k) \cdot s_d(k), & \text{ otherwise}. 
\end{cases}$
\STATE \textbf{return} $\algname(\boldsymbol{g}^1, \dots, \boldsymbol{g}^N) =\frac{\sum_{k \in [N]} s(k)\cdot\boldsymbol{g}^k}{\sum_{i \in [N]}s(k)}$
\end{algorithmic}
\end{algorithm}

\algname is based on two main design principles:
 \begin{enumerate}[(i)]  
    \item Colluding Byzantine clients can bypass similarity-based Byzantine-robust aggregation methods by artificially creating highly similar malicious local updates. A robust aggregation method that successfully defends against such attacks should penalize over-proportional proximity.
    \item Under a non-IID data distribution in particular, a single honest gradient is not representative of the entire set of honest clients $\mathcal{H}$. Therefore, a successful aggregation scheme should output a mix of honest updates.
\end{enumerate}
To implement the first principle, \algname excludes the $f-1$ nearest and $f$ farthest neighbors in the computation of the proximity score, cf. \cref{eq:score1}. The farthest $f$ neighbors are excluded to mitigate the influence of up to $f$ malicious clients, under the assumption that adversarial gradients are intentionally distant from those of honest clients. The nearest $f-1$ distances are disregarded to prevent colluding clients from being assigned disproportionately high proximity scores.
The dissimilarity score further penalizes suspiciously similar gradients by integrating the coefficient of variance among the neighboring clients into the trust score \cref{eq:score2}.
To satisfy the second principle, \algname linearly mixes highly scored gradients to overcome the negative impact of non-IID distributions. At the same time, it uses a relative threshold to overcome the vulnerabilities of purely linear aggregations~\cite{krum}.
\paragraph*{Complexity analysis} 
\algname algorithm starts by computing the pairwise distance matrix in order to derive proximity and dissimilarity scores. Constructing this matrix involves $\mathcal{O}(N^2d)$ operations. Then, for each client $k$, the proximity score $s_p(k)$ is calculated by sorting and selecting closest updates from the window frame $[f, N-f-1]$. Using Quickselect, this operation requires $\mathcal{O}(N)$ expected complexity. The second score $s_d(k)$ for client $k$ requires forming the set $\mathcal{N}_f(k)$ of $f$ nearest neighbors, which again requires $\mathcal{O}(N)$ operations in expectation with Quickselect. Computing the mean and the standard deviation of the set $\mathcal{N}_f(k)$ incurs an additional cost of $\mathcal{O}(fd)$. After performing these operations for all $N$ clients, the scores are aggregated with complexity $\mathcal{O}(N)$. Finally, the $f$ smallest scores are identified and set to $0$, with an additional complexity of $\mathcal{O}(N)$ in expectation. Overall, the algorithm has expected complexity of $\mathcal{O}(N^2d)+\mathcal{O}(Nfd)$. Noting that $f < N/2$, 
the total expected complexity of \algname is $\mathcal{O}(N^2d)$.

State-of-the-art defense methods that leverage pairwise distance information have expected time complexity $\mathcal{O}(N^2d)$, e.g., Krum, Clustering Methods, etc. While some methods, e.g., Coordinate-wise Trimmed Mean, have expected complexity of $\mathcal{O}(Nd)$, it is shown that combining them with pre-aggregation methods, e.g., NNM, improves their robustness at the expense of an increased complexity.

\section{Experimental Evaluation}

\subsection{Experimental Setup}
\label{sec:setup}

We perform extensive experiments to compare the performance of \algname to state-of-the-art robust aggregation methods for various settings and under different attacks.
The experiments consider an FL environment with $N \in \{10,20,30,50,100\}$ clients and $\frac{f}{N}\in\{0.1,0.2,0.3\}$. The task is an image classification on the CIFAR-10 \cite{cifar10} and FEMNIST \cite{leaf} datasets that have $10$ and $62$ classes of images, respectively. Local training is performed with a batch size $b=64$. For non-IID setting of CIFAR-10 dataset, the data distribution of each class $j \in [10]$ is a Dirichlet distribution\cite{dirichlet}, i.e., $\boldsymbol{p}_{j} \sim \text{Dir}_N(\alpha)$, with parameter $\alpha=0.1$ (strongly non-IID) is chosen. The source of non-IID distribution, thus, becomes label distribution skew. However, for the FEMNIST dataset, the source of non-IID distribution is the feature distribution skew. FEMNIST consists of $3500$ different users with varying numbers of personal handwriting samples. We sample the user datasets using the LEAF \cite{leaf} framework such that the minimum number of samples per user is $350$. The training-test dataset ratio is chosen as $0.9/0.1$. Training samples are used during the local training of the model, whereas test samples from selected clients are used to evaluate the performance of the global model. 

For CIFAR-10, we use the CNN model from~\cite{vitalrole} that has $d=1,310,922$ parameters. The model is trained using a negative log-likelihood loss combined with an $\ell_2$-regularization term with a regularization factor of  $10^{-2}$. For FEMNIST, we use a CNN model with $d=1,690,046$ and the cross-entropy loss function. The number of local iterations performed by each client is chosen to be either $E=1$ (known as \texttt{FedSGD}~\cite{mcmahan}) or $E=10$ (known as \texttt{FedAVG}~\cite{mcmahan}).
The FL algorithm is run for $T=2000$ rounds for \texttt{FedSGD} and $T\in\{800, 1000\}$ rounds for \texttt{FedAVG}.
When local momentum is applied, we use $\beta=0.9$. The learning rate in round $t$ is
$$
\gamma_t = \begin{cases}
 0.05 &\quad \text{if } t\leq \frac{2}{3}T,\\
 0.005 &\quad \text{else.} 
 \end{cases}
$$

\paragraph{Attacks for Evaluation}

 Six different attack scenarios are performed: No Attack, A little is enough (ALIE) \cite{alie}, Fall of Empires (FOE) \cite{foe} with $\epsilon=0.1$ and $\epsilon=100$, Label-Flip (LF) \cite{lfsf}, and Sign-Flip (SF)\cite{lfsf}. For ALIE and FOE, parameters ($\epsilon$ and $z$, respectively) are further optimized through linear search (similar to \cite{fixbymix,manipulatingthebyzantine}), where the parameter giving the highest $\ell_2$ distance of the resulting aggregation to the average of the honest clients' gradients is chosen. We use $\boldsymbol{\mu}_\mathcal{H}$ and $\boldsymbol{\sigma}_\mathcal{H}$ to denote the mean and the standard deviation of the honest client gradients, respectively. A Byzantine client $k$ sends a manipulated gradient $\boldsymbol{\tilde{g}}^k$ instead of the honest value $\boldsymbol{g}^k$, such that: 
\begin{itemize}
    \item ALIE($z$): $\boldsymbol{\tilde{g}}^k =  \boldsymbol{\mu}_\mathcal{H} - z^* \boldsymbol{\sigma}_\mathcal{H}$, where $z^* \in \{0.25z, 0.5z, \dots, c^* z\}$ where $c^*$ is the largest multiple of $0.25$ such that $c^* z \leq 2$.
    \item FOE($\epsilon$): $\boldsymbol{\tilde{g}}^k =  -\epsilon^* \boldsymbol{\mu}_\mathcal{H}$, where $\epsilon^* \in \{0.1\epsilon, 0.2\epsilon, \dots, \epsilon\}$.
    \item SF: $\boldsymbol{\tilde{g}}^k = - \boldsymbol{g}^k$, where the Byzantine client sends the negative of the gradient computed on its local data.
    \item LF: $\boldsymbol{\tilde{g}}^k$ computed on the data with flipped labels; $y' = 9-y$ for CIFAR-10, $y'=61-y$ for FEMNIST.\\
\end{itemize}
\begin{table*}[tb]
\renewcommand{\arraystretch}{1.3}
\caption{Final accuracies for CIFAR-10 dataset, $N=10$, $f=3$, $T=2000$ for \texttt{FedSGD} and $T=1000$ for \texttt{FedAVG}.Worst-case accuracies for each defense method are shown in bold. }
\label{tab:bigtable}
    \centering
    \resizebox{\textwidth}{!}{
    \begin{tabular}
    {||c||c||c|p{1cm}| p{1cm}| p{1cm}| p{1cm}| p{1cm}|p{1cm}|| p{1cm}| p{1cm}|p{1cm}| p{1cm}| p{1cm}|p{1cm}||}
        \hline
        Local Momentum & Algorithm & Defense & \multicolumn{6}{|c||}{IID Data} & \multicolumn{6}{|c||}{Non-IID Data (Dirichlet $\alpha = 0.1$)}\\
        \hline
        \hline
        \multirow{15}{*}{\rotatebox{90}{$\beta=0$}}
            & \multirow{8}{*}{\rotatebox{90}{\texttt{FedSGD}\quad ($E=1$)}} & & \textbf{No \quad Attack} & \textbf{ALIE} & \textbf{FOE} \quad\quad $\epsilon=0.1$  & \textbf{FOE} \quad\quad$\epsilon=100$ & \textbf{LF} & \textbf{SF}  & \textbf{No \quad Attack} & \textbf{ALIE} & \textbf{FOE \quad\quad$\epsilon=0.1$} & \textbf{FOE \quad\quad $\epsilon=100$} & \textbf{LF} & \textbf{SF}\\ 
            \hhline{~~-------------} 
            & & No Defense & \cellcolor{Salmon!5} 78.31 &\cellcolor{Salmon!5}  10.11 &\cellcolor{Salmon!5} 75.07 &\cellcolor{Salmon!5} \textbf{10.00} &\cellcolor{Salmon!5} 73.59  &\cellcolor{Salmon!5}66.67 &\cellcolor{YellowGreen!5}59.34  &\cellcolor{YellowGreen!5}  \textbf{10.00}&\cellcolor{YellowGreen!5} 54.07 &\cellcolor{YellowGreen!5}  \textbf{10.00}&\cellcolor{YellowGreen!5} 44.57 &\cellcolor{YellowGreen!5} \textbf{10.00} \\
            & & \texttt{NNM + Median} & \cellcolor{Salmon!5} 78.54 &\cellcolor{Salmon!5} \textbf{10.00} &\cellcolor{Salmon!5}72.25  &\cellcolor{Salmon!5}  77.52 &\cellcolor{Salmon!5}77.99  &\cellcolor{Salmon!5} 71.70 &\cellcolor{YellowGreen!5} 60.02  &\cellcolor{YellowGreen!5}  \textbf{10.00}&\cellcolor{YellowGreen!5} 23.40 &\cellcolor{YellowGreen!5} 57.05 &\cellcolor{YellowGreen!5}44.45  &\cellcolor{YellowGreen!5} 54.52\\
            & & \texttt{NNM + TrimmedMean} & \cellcolor{Salmon!5} 78.14 &\cellcolor{Salmon!5} \textbf{10.00}  &\cellcolor{Salmon!5}  72.18 &\cellcolor{Salmon!5}77.70  &\cellcolor{Salmon!5}77.52  &\cellcolor{Salmon!5} 71.84 &\cellcolor{YellowGreen!5}60.43  &\cellcolor{YellowGreen!5}  10.00&\cellcolor{YellowGreen!5} \textbf{9.85} &\cellcolor{YellowGreen!5} 57.67 &\cellcolor{YellowGreen!5}  44.68&\cellcolor{YellowGreen!5}54.70 \\
            & & \texttt{NNM + GeoMed}& \cellcolor{Salmon!5} 78.14 &\cellcolor{Salmon!5} \textbf{10.00} &\cellcolor{Salmon!5} 72.29 &\cellcolor{Salmon!5}  76.82&\cellcolor{Salmon!5} 77.42 &\cellcolor{Salmon!5} 71.37 &\cellcolor{YellowGreen!5} 60.43  &\cellcolor{YellowGreen!5}  \textbf{10.00}&\cellcolor{YellowGreen!5} 22.12 &\cellcolor{YellowGreen!5}57.11  &\cellcolor{YellowGreen!5} 43.41 &\cellcolor{YellowGreen!5}55.51 \\
            & & \texttt{NNM + Krum}& \cellcolor{Salmon!5} 78.29 &\cellcolor{Salmon!5}  \textbf{10.00}&\cellcolor{Salmon!5} 72.22 &\cellcolor{Salmon!5}  77.42 &\cellcolor{Salmon!5}77.50  &\cellcolor{Salmon!5}72.30  &\cellcolor{YellowGreen!5} 59.56 &\cellcolor{YellowGreen!5}  \textbf{10.00}&\cellcolor{YellowGreen!5} 16.60 &\cellcolor{YellowGreen!5}  57.28&\cellcolor{YellowGreen!5} 39.04  &\cellcolor{YellowGreen!5} 54.24\\
            & & \texttt{NNM + CClip}& \cellcolor{Salmon!5} 78.39 &\cellcolor{Salmon!5}  \textbf{10.00}&\cellcolor{Salmon!5} 72.36 &\cellcolor{Salmon!5} \textbf{10.00} &\cellcolor{Salmon!5} 75.54 &\cellcolor{Salmon!5} 70.66 &\cellcolor{YellowGreen!5} 60.28 &\cellcolor{YellowGreen!5}  12.28&\cellcolor{YellowGreen!5} 48.91  &\cellcolor{YellowGreen!5}  \textbf{9.97}&\cellcolor{YellowGreen!5} 45.86 &\cellcolor{YellowGreen!5}50.58 \\
            \hhline{~~-------------} 
            & & \algname (ours) & \cellcolor{Salmon!5} 78.45 &\cellcolor{Salmon!5} 77.35 &\cellcolor{Salmon!5} 77.66 &\cellcolor{Salmon!5}77.68  &\cellcolor{Salmon!5} 77.66 &\cellcolor{Salmon!5} \textbf{67.75} &\cellcolor{YellowGreen!5}  58.40 &\cellcolor{YellowGreen!5}55.94  &\cellcolor{YellowGreen!5}  55.79&\cellcolor{YellowGreen!5}57.79  &\cellcolor{YellowGreen!5}  \textbf{45.24}&\cellcolor{YellowGreen!5} 52.66\\
            \hhline{~--------------} 
            &\multirow{7}{*}{\rotatebox{90}{\texttt{FedAVG}\quad ($E=10$)}} 
            & No Defense & \cellcolor{Maroon!5} 85.94&\cellcolor{Maroon!5} 50.76 &\cellcolor{Maroon!5} 83.87 &\cellcolor{Maroon!5}\textbf{10.00}  &\cellcolor{Maroon!5} 66.15 &\cellcolor{Maroon!5} 73.36 &\cellcolor{YellowGreen!10}75.25  &\cellcolor{YellowGreen!10}  14.48&\cellcolor{YellowGreen!10} 68.17 &\cellcolor{YellowGreen!10}  \textbf{10.00}&\cellcolor{YellowGreen!10} 58.83 &\cellcolor{YellowGreen!10}\textbf{10.00} \\
            & & \texttt{NNM + Median} & \cellcolor{Maroon!5}85.44 &\cellcolor{Maroon!5}  \textbf{57.80} &\cellcolor{Maroon!5}82.03  &\cellcolor{Maroon!5} 84.56 &\cellcolor{Maroon!5} 84.68 &\cellcolor{Maroon!5} 83.90 &\cellcolor{YellowGreen!10}77.07 &\cellcolor{YellowGreen!10}  \textbf{11.69}&\cellcolor{YellowGreen!10} 53.35 &\cellcolor{YellowGreen!10}  68.44&\cellcolor{YellowGreen!10} 57.97 &\cellcolor{YellowGreen!10}67.75 \\
            & & \texttt{NNM + TrimmedMean} & \cellcolor{Maroon!5} 85.78 &\cellcolor{Maroon!5} \textbf{57.65} &\cellcolor{Maroon!5}  82.12&\cellcolor{Maroon!5} 84.41 &\cellcolor{Maroon!5} 85.16 &\cellcolor{Maroon!5}83.46  &\cellcolor{YellowGreen!10} 77.01 &\cellcolor{YellowGreen!10}\textbf{10.33}  &\cellcolor{YellowGreen!10}  53.46&\cellcolor{YellowGreen!10} 69.41 &\cellcolor{YellowGreen!10}  57.96&\cellcolor{YellowGreen!10} 69.32\\
            & & \texttt{NNM + GeoMed}& \cellcolor{Maroon!5} 85.78&\cellcolor{Maroon!5}  \textbf{57.45}&\cellcolor{Maroon!5} 81.84 &\cellcolor{Maroon!5}  84.47&\cellcolor{Maroon!5} 85.07 &\cellcolor{Maroon!5}  84.07&\cellcolor{YellowGreen!10} 76.73 &\cellcolor{YellowGreen!10}  \textbf{20.97}&\cellcolor{YellowGreen!10}54.98 &\cellcolor{YellowGreen!10}  69.46&\cellcolor{YellowGreen!10} 56.02 &\cellcolor{YellowGreen!10}69.37 \\
            & & \texttt{NNM + Krum}& \cellcolor{Maroon!5} 85.78 &\cellcolor{Maroon!5}  \textbf{54.34}&\cellcolor{Maroon!5} 82.11 &\cellcolor{Maroon!5} 84.97 &\cellcolor{Maroon!5} 84.93 &\cellcolor{Maroon!5} 84.19  &\cellcolor{YellowGreen!10}74.05  &\cellcolor{YellowGreen!10}  \textbf{14.06}&\cellcolor{YellowGreen!10} 54.03 &\cellcolor{YellowGreen!10} 68.27 &\cellcolor{YellowGreen!10} 57.78 &\cellcolor{YellowGreen!10} 67.84\\
            & & \texttt{NNM + CClip}& \cellcolor{Maroon!5}86.08 &\cellcolor{Maroon!5}\textbf{57.80} &\cellcolor{Maroon!5}  82.05&\cellcolor{Maroon!5} 67.93 &\cellcolor{Maroon!5} 85.03 &\cellcolor{Maroon!5} 83.74 &\cellcolor{YellowGreen!10}  76.55&\cellcolor{YellowGreen!10} \textbf{23.74} &\cellcolor{YellowGreen!10} 59.94  &\cellcolor{YellowGreen!10}47.88  &\cellcolor{YellowGreen!10} 61.82 &\cellcolor{YellowGreen!10} 65.89\\
             \hhline{~~-------------} 
            & & \algname (ours) & \cellcolor{Maroon!5} 85.73 &\cellcolor{Maroon!5}  85.04&\cellcolor{Maroon!5} 84.32 &\cellcolor{Maroon!5} 84.43 &\cellcolor{Maroon!5} 84.83 &\cellcolor{Maroon!5} \textbf{83.44} &\cellcolor{YellowGreen!10}74.72  &\cellcolor{YellowGreen!10}  66.14&\cellcolor{YellowGreen!10} 68.30 &\cellcolor{YellowGreen!10}  68.11&\cellcolor{YellowGreen!10}  62.69&\cellcolor{YellowGreen!10} \textbf{61.46}\\
            \hline\hline
            \multirow{15}{*}{\rotatebox{90}{$\beta=0.9$}}
            & \multirow{8}{*}{\rotatebox{90}{\texttt{FedSGD}\quad ($E=1$)}} & & \textbf{No \quad Attack} & \textbf{ALIE} & \textbf{FOE}\quad\quad $\epsilon=0.1$ &\textbf{FOE} \quad\quad$\epsilon=100$ & \textbf{LF} & \textbf{SF}  & \textbf{No \quad Attack} & \textbf{ALIE} & \textbf{FOE \quad\quad $\epsilon=0.1$} & \textbf{FOE} \quad\quad $\epsilon=100$  & \textbf{LF} & \textbf{SF}\\ 
            \hhline{~~-------------} 
            & &No Defense &\cellcolor{Purple!5} 78.30&\cellcolor{Purple!5} 45.90 & \cellcolor{Purple!5} 74.00 & \cellcolor{Purple!5} \textbf{10.00} & \cellcolor{Purple!5} 72.18 & \cellcolor{Purple!5} 73.97& \cellcolor{SkyBlue!5} 58.25 & \cellcolor{SkyBlue!5} 10.20 & \cellcolor{SkyBlue!5}54.82 & \cellcolor{SkyBlue!5} \textbf{10.00}&\cellcolor{SkyBlue!5} 42.06 &\cellcolor{SkyBlue!5} 52.52 \\
            & & \texttt{NNM + Median} &\cellcolor{Purple!5} 78.11&\cellcolor{Purple!5} \textbf{51.91} & \cellcolor{Purple!5} 71.99 & \cellcolor{Purple!5} 77.03& \cellcolor{Purple!5} 76.55& \cellcolor{Purple!5} 72.93 &\cellcolor{SkyBlue!5}57.89 &  \cellcolor{SkyBlue!5}\textbf{10.26} & \cellcolor{SkyBlue!5} 26.89 & \cellcolor{SkyBlue!5} 46.54 &\cellcolor{SkyBlue!5}  49.57&\cellcolor{SkyBlue!5} 43.46 \\
            & & \texttt{NNM + TrimmedMean} &\cellcolor{Purple!5}78.14 &\cellcolor{Purple!5} \textbf{51.35} & \cellcolor{Purple!5} 71.94 & \cellcolor{Purple!5}77.05 & \cellcolor{Purple!5} 76.53 & \cellcolor{Purple!5}72.80 &\cellcolor{SkyBlue!5}56.92 & \cellcolor{SkyBlue!5} \textbf{10.02} & \cellcolor{SkyBlue!5}15.01 & \cellcolor{SkyBlue!5}47.45 &\cellcolor{SkyBlue!5}  47.21&\cellcolor{SkyBlue!5} 49.07\\
            & & \texttt{NNM + GeoMed} &\cellcolor{Purple!5} 78.14&\cellcolor{Purple!5} \textbf{51.41} & \cellcolor{Purple!5}72.05  & \cellcolor{Purple!5}76.27 & \cellcolor{Purple!5} 75.91& \cellcolor{Purple!5} 73.04&\cellcolor{SkyBlue!5} 56.92 & \cellcolor{SkyBlue!5} \textbf{9.97} & \cellcolor{SkyBlue!5} 16.91 & \cellcolor{SkyBlue!5} 44.27 &\cellcolor{SkyBlue!5} 48.22 &\cellcolor{SkyBlue!5} 48.90\\
            & & \texttt{NNM + Krum} &\cellcolor{Purple!5}78.32 &\cellcolor{Purple!5} \textbf{48.47}  & \cellcolor{Purple!5} 71.52 & \cellcolor{Purple!5} 77.30& \cellcolor{Purple!5} 76.59& \cellcolor{Purple!5} 73.37 &\cellcolor{SkyBlue!5} 58.12 & \cellcolor{SkyBlue!5}\textbf{10.02} & \cellcolor{SkyBlue!5} 16.66& \cellcolor{SkyBlue!5}48.73 &\cellcolor{SkyBlue!5}  41.61&\cellcolor{SkyBlue!5}32.91 \\
            & & \texttt{NNM + CClip} &\cellcolor{Purple!5}78.26 &\cellcolor{Purple!5} 54.75 & \cellcolor{Purple!5} 71.77 & \cellcolor{Purple!5} \textbf{10.00}& \cellcolor{Purple!5} 74.62& \cellcolor{Purple!5} 73.01 &\cellcolor{SkyBlue!5} 57.22 & \cellcolor{SkyBlue!5} 11.71 & \cellcolor{SkyBlue!5}47.42 & \cellcolor{SkyBlue!5} \textbf{9.96} &\cellcolor{SkyBlue!5} 44.25 &\cellcolor{SkyBlue!5} 46.68\\
            \hhline{~~-------------} 
            & & \algname (ours) &\cellcolor{Purple!5} 78.10&\cellcolor{Purple!5} 76.53 & \cellcolor{Purple!5} 76.73 & \cellcolor{Purple!5} 76.82 & \cellcolor{Purple!5} 77.06 & \cellcolor{Purple!5} \textbf{71.39}&\cellcolor{SkyBlue!5} 58.28& \cellcolor{SkyBlue!5}50.20 & \cellcolor{SkyBlue!5}52.86 & \cellcolor{SkyBlue!5}56.37 &\cellcolor{SkyBlue!5}  45.08&\cellcolor{SkyBlue!5}\textbf{38.62 }\\

            \hhline{~--------------} 
            &\multirow{7}{*}{\rotatebox{90}{\texttt{FedAVG}\quad ($E=10$)}} 
            & No Defense & \cellcolor{Purple!8}85.12 &\cellcolor{Purple!8} 78.61& \cellcolor{Purple!8} 83.51 &\cellcolor{Purple!8} \textbf{10.00}  &\cellcolor{Purple!8} 74.78 &\cellcolor{Purple!8}82.81  &\cellcolor{MidnightBlue!5} 69.88  & \cellcolor{MidnightBlue!5} 23.59 &  \cellcolor{MidnightBlue!5} 68.54& \cellcolor{MidnightBlue!5}\textbf{10.00} & \cellcolor{MidnightBlue!5} 55.24 & \cellcolor{MidnightBlue!5} 63.96\\
            & & \texttt{NNM + Median}& \cellcolor{Purple!8}85.24 &\cellcolor{Purple!8}\textbf{80.57}& \cellcolor{Purple!8}80.61 &\cellcolor{Purple!8}  84.59&\cellcolor{Purple!8} 84.68 &\cellcolor{Purple!8}  81.72 &\cellcolor{MidnightBlue!5} 76.92 & \cellcolor{MidnightBlue!5} \textbf{15.89} &  \cellcolor{MidnightBlue!5} 59.18& \cellcolor{MidnightBlue!5}67.22 & \cellcolor{MidnightBlue!5}61.44 & \cellcolor{MidnightBlue!5} 58.70\\
            & & \texttt{NNM + TrimmedMean}& \cellcolor{Purple!8} 85.00&\cellcolor{Purple!8}80.93 & \cellcolor{Purple!8}\textbf{80.56} &\cellcolor{Purple!8} 84.30 &\cellcolor{Purple!8} 84.52  &\cellcolor{Purple!8} 81.47  &\cellcolor{MidnightBlue!5} 77.01  & \cellcolor{MidnightBlue!5} \textbf{11.27} &  \cellcolor{MidnightBlue!5} 59.55& \cellcolor{MidnightBlue!5} 68.29 & \cellcolor{MidnightBlue!5} 63.47& \cellcolor{MidnightBlue!5} 59.22\\
            & & \texttt{NNM + GeoMed} & \cellcolor{Purple!8} 85.00 &\cellcolor{Purple!8} \textbf{80.75} & \cellcolor{Purple!8}81.20 &\cellcolor{Purple!8} 84.58 &\cellcolor{Purple!8} 84.73  &\cellcolor{Purple!8} 81.71 &\cellcolor{MidnightBlue!5} 76.66  & \cellcolor{MidnightBlue!5} \textbf{12.07} &  \cellcolor{MidnightBlue!5} 59.27& \cellcolor{MidnightBlue!5} 68.19 & \cellcolor{MidnightBlue!5} 68.92 & \cellcolor{MidnightBlue!5} 51.29\\
            & & \texttt{NNM + Krum}& \cellcolor{Purple!8} 84.94 &\cellcolor{Purple!8}\textbf{79.89} & \cellcolor{Purple!8} 80.22 &\cellcolor{Purple!8} 84.29 &\cellcolor{Purple!8} 84.28 &\cellcolor{Purple!8} 80.36 &\cellcolor{MidnightBlue!5} 70.07 & \cellcolor{MidnightBlue!5} \textbf{18.77}&  \cellcolor{MidnightBlue!5} 59.33& \cellcolor{MidnightBlue!5}62.95 & \cellcolor{MidnightBlue!5} 52.31& \cellcolor{MidnightBlue!5} 39.56 \\
            & & \texttt{NNM + CClip}& \cellcolor{Purple!8} 84.83 &\cellcolor{Purple!8} 80.48& \cellcolor{Purple!8}81.34 &\cellcolor{Purple!8} \textbf{37.79} &\cellcolor{Purple!8} 83.83  &\cellcolor{Purple!8} 82.42 &\cellcolor{MidnightBlue!5} 76.54 & \cellcolor{MidnightBlue!5} \textbf{25.45} &  \cellcolor{MidnightBlue!5} 65.88& \cellcolor{MidnightBlue!5} 36.51 & \cellcolor{MidnightBlue!5} 69.11& \cellcolor{MidnightBlue!5} 61.91\\
            \hhline{~~-------------} 
            & & \algname (ours) & \cellcolor{Purple!8} 85.00 &\cellcolor{Purple!8} 84.44 & \cellcolor{Purple!8}84.95 &\cellcolor{Purple!8} 84.87 &\cellcolor{Purple!8} 84.00 &\cellcolor{Purple!8} \textbf{80.10} &\cellcolor{MidnightBlue!5} 66.15 & \cellcolor{MidnightBlue!5} 57.16&  \cellcolor{MidnightBlue!5} 64.88& \cellcolor{MidnightBlue!5}66.08 & \cellcolor{MidnightBlue!5} 57.86& \cellcolor{MidnightBlue!5} \textbf{46.65}\\
            \hline
    \end{tabular}
    }
\end{table*}

\begin{table}[tb]
\renewcommand{\arraystretch}{1.7}
\caption{Final accuracies for FEMNIST dataset, $N=20$, $f=6$, $T=800$. Worst-case accuracies for each defense method are shown in bold. All experiments are run with three random seeds.}\label{tab:femnistt}
    \centering
    \resizebox{\columnwidth}{!}{
    
\begin{tabular}
    {||p{1.2cm}||c|p{1.5cm}| p{1.85cm}| p{1.6cm}| p{1.5cm}| p{1.5cm}|p{1.8cm}||}
        \hline
        Momentum, Algorithm  & Defense & \multicolumn{6}{|c||}{non-IID Data} \\
        \hline
        \hline
            \multirow{9}{*}{\rotatebox{90}{\parbox{3cm}{\centering$\beta=0$, \vspace{0.3cm} \\ \texttt{FedAVG} ($E=10$)}}} & & \textbf{No \quad Attack} & \textbf{ALIE} & \textbf{FOE} \quad\quad $\epsilon=0.1$  & \textbf{FOE} \quad\quad$\epsilon=100$ & \textbf{LF} & \textbf{SF} \\ 
            \hhline{~-------} 
            & No Defense & \cellcolor{YellowGreen!10} $87.89 \pm 0.45$ &\cellcolor{YellowGreen!10} $5.38 \pm 1.13$ & \cellcolor{YellowGreen!10} $83.90 \pm 0.66$ &\cellcolor{YellowGreen!10}  \boldmath{$3.78\pm0.00$} &\cellcolor{YellowGreen!10}  $67.76\pm0.12$ &\cellcolor{YellowGreen!10}\boldmath{$3.78\pm0.00$} \\ 
            & \texttt{NNM + Median}& \cellcolor{YellowGreen!10}  $88.27\pm0.54$ & \cellcolor{YellowGreen!10} \boldmath{$5.38\pm1.13$} &\cellcolor{YellowGreen!10} $71.63\pm1.42$& \cellcolor{YellowGreen!10} $80.50\pm0.77$ &\cellcolor{YellowGreen!10} $83.86\pm0.54$  &\cellcolor{YellowGreen!10}  $78.98\pm2.07$ \\
            &  \texttt{NNM + TrimmedMean}& \cellcolor{YellowGreen!10} $88.06 \pm 0.59$ &\cellcolor{YellowGreen!10}\boldmath{$6.18 \pm 0.00$} & \cellcolor{YellowGreen!10}$70.91 \pm 1.24$  &\cellcolor{YellowGreen!10}  $81.13 \pm 0.98$ &\cellcolor{YellowGreen!10} $83.48 \pm 0.31$ &\cellcolor{YellowGreen!10} $ 80.03 \pm 1.31$\\
            &  \texttt{NNM + GeoMed} & \cellcolor{YellowGreen!10} $88.06\pm 0.59$ &\cellcolor{YellowGreen!10}\boldmath{$5.38\pm 1.13$} & \cellcolor{YellowGreen!10} $69.61\pm 1.03$  &\cellcolor{YellowGreen!10} $79.91\pm 0.80$ &\cellcolor{YellowGreen!10} $83.77\pm 0.49$  &\cellcolor{YellowGreen!10} $79.53\pm 2.01$ \\
            &  \texttt{NNM + Krum}& \cellcolor{YellowGreen!10} $88.23\pm 0.52$ &\cellcolor{YellowGreen!10} \boldmath{$5.34\pm 1.10$} & \cellcolor{YellowGreen!10}$26.19\pm 31.68$ &\cellcolor{YellowGreen!10} $81.80\pm 0.57$ &\cellcolor{YellowGreen!10} $83.69\pm 0.52 $&\cellcolor{YellowGreen!10} $79.11\pm 1.64$ \\
            &  \texttt{NNM + CClip}& \cellcolor{YellowGreen!10}$88.31\pm 0.36$ &\cellcolor{YellowGreen!10} $79.82\pm 1.04$ & \cellcolor{YellowGreen!10}$77.47\pm 0.84$  &\cellcolor{YellowGreen!10} \boldmath{$5.63\pm 0.42$}   &\cellcolor{YellowGreen!10} $78.44\pm 0.18$ &\cellcolor{YellowGreen!10} $69.02\pm 1.61$\\
            \hhline{~-------} 
            &  \algname (ours) &\cellcolor{YellowGreen!10} $87.89\pm 0.68$ & \cellcolor{YellowGreen!10} $82.85\pm 0.37$  &\cellcolor{YellowGreen!10} $82.35\pm 0.47$ & \cellcolor{YellowGreen!10}  $83.73\pm 0.72$ &\cellcolor{YellowGreen!10} $83.90\pm 0.52$  &\cellcolor{YellowGreen!10}  \boldmath{$77.47\pm 0.12$}  \\
           
            \hline\hline
            \multirow{8}{*}{\rotatebox{90}{\parbox{3cm}{\centering$\beta=0.9$, \vspace{0.3cm} \\ \texttt{FedAVG} ($E=10$)}}} 
            & No Defense & \cellcolor{MidnightBlue!5} $87.68\pm 0.46$ &\cellcolor{MidnightBlue!5}$12.65\pm 9.42$ & \cellcolor{MidnightBlue!5} $83.82\pm 0.51$ &\cellcolor{MidnightBlue!5} \boldmath{$3.78\pm 0.00$}  &\cellcolor{MidnightBlue!5} $66.67\pm 0.99$  &\cellcolor{MidnightBlue!5} $81.04\pm 1.45$ \\
            &  \texttt{NNM + Median} & \cellcolor{MidnightBlue!5} $87.98\pm 0.49$ &\cellcolor{MidnightBlue!5} \boldmath{$6.18\pm 0.00$} & \cellcolor{MidnightBlue!5} $47.54\pm 30.95$ &\cellcolor{MidnightBlue!5} $79.57\pm 0.27$  &\cellcolor{MidnightBlue!5}  $83.98\pm 0.18$ &\cellcolor{MidnightBlue!5} $67.93\pm 0.39$ \\
            &  \texttt{NNM + TrimmedMean}& \cellcolor{MidnightBlue!5} $87.68\pm 0.36$ &\cellcolor{MidnightBlue!5} \boldmath{$5.38\pm 1.13$} & \cellcolor{MidnightBlue!5} $70.32\pm 0.42$ &\cellcolor{MidnightBlue!5} $7$$9.57\pm 0.57$  &\cellcolor{MidnightBlue!5} $83.90\pm 0.57$ &\cellcolor{MidnightBlue!5} $66.37\pm 2.01$ \\
            &  \texttt{NNM + GeoMed} & \cellcolor{MidnightBlue!5} $87.68\pm 0.36$ &\cellcolor{MidnightBlue!5} \boldmath{$21.86\pm22.17$} & \cellcolor{MidnightBlue!5} $69.02\pm 1.63$ &\cellcolor{MidnightBlue!5} $80.33\pm 0.74$  &\cellcolor{MidnightBlue!5}  $84.03\pm 0.36$ &\cellcolor{MidnightBlue!5} $68.73\pm 0.90$ \\
            &  \texttt{NNM + Krum}& \cellcolor{MidnightBlue!5} $87.77\pm 0.54$ &\cellcolor{MidnightBlue!5} \boldmath{$6.14\pm 0.06$} & \cellcolor{MidnightBlue!5} $47.79\pm 31.13$ &\cellcolor{MidnightBlue!5} $78.90\pm 1.06$  &\cellcolor{MidnightBlue!5}  $83.98\pm 0.31$ &\cellcolor{MidnightBlue!5} $66.50\pm 1.02$ \\
            & \texttt{NNM + CClip}& \cellcolor{MidnightBlue!5} $87.60\pm 0.39$ &\cellcolor{MidnightBlue!5} $51.07\pm 32.18$ & \cellcolor{MidnightBlue!5} $78.39\pm 0.73$ &\cellcolor{MidnightBlue!5} \boldmath{$1.77\pm 0.00$}  &\cellcolor{MidnightBlue!5} $78.94\pm 0.21$ &\cellcolor{MidnightBlue!5} $74.74\pm 1.00$ \\
            \hhline{~-------}  
            &  \algname (ours) & \cellcolor{MidnightBlue!5} $87.77\pm 0.64$ &\cellcolor{MidnightBlue!5} $82.72\pm 0.51$ & \cellcolor{MidnightBlue!5} $82.56\pm 0.67$ &\cellcolor{MidnightBlue!5} $83.52\pm 0.51$ &\cellcolor{MidnightBlue!5}  $83.69\pm 0.76$ &\cellcolor{MidnightBlue!5} \boldmath{$68.43\pm 1.80$} \\
           
            \hline\hline
    \end{tabular}

    }
\end{table}
\begin{table}[tb]
\renewcommand{\arraystretch}{1.7}
\caption{Final accuracies for FEMNIST dataset, $N=100$, $f=30$, $T=800$. Worst-case accuracies for each defense method are shown in bold. All experiments are run with three random seeds.}\label{tab:femnistt100}
    \centering
    \resizebox{\columnwidth}{!}{
    
\begin{tabular}
    {||p{1.2cm}||c| p{1.85cm}| p{1.85cm}| p{1.6cm}| p{1.5cm}|p{1.8cm}||}
        \hline
        Momentum, Algorithm  & Defense & \multicolumn{5}{|c||}{non-IID Data} \\
        \hline
        \hline
            \multirow{9}{*}{\rotatebox{90}{\parbox{3cm}{\centering$\beta=0.9$, \vspace{0.3cm} \\ \texttt{FedAVG} ($E=10$)}}} & & \textbf{ALIE} & \textbf{FOE} \quad\quad\quad $\epsilon=0.1$  & \textbf{FOE} \quad\quad$\epsilon=100$ & \textbf{LF} & \textbf{SF} \\ 
            \hhline{~------} 
            & No Defense  &\cellcolor{MidnightBlue!5} $4.03\pm1.21$ & \cellcolor{MidnightBlue!5} $87.44\pm0.26$  &\cellcolor{MidnightBlue!5} \boldmath{$3.18\pm0.00$}  &\cellcolor{MidnightBlue!5} $75.99\pm0.69$  &\cellcolor{MidnightBlue!5} $87.32\pm0.16$ \\
            &  \texttt{NNM + Median}  &\cellcolor{MidnightBlue!5} \boldmath{$4.03\pm1.21$} & \cellcolor{MidnightBlue!5} $28.85 \pm 36.31$ & \cellcolor{MidnightBlue!5}  $58.93\pm37.10$    &\cellcolor{MidnightBlue!5}  $88.17\pm0.15$ &\cellcolor{MidnightBlue!5} $80.65\pm0.15$ \\
            &  \texttt{NNM + TrimmedMean}&\cellcolor{MidnightBlue!5} \boldmath{$5.74\pm0.00$} & \cellcolor{MidnightBlue!5} $54.45\pm36.26$ &  \cellcolor{MidnightBlue!5} $30.07\pm38.04$  & \cellcolor{MidnightBlue!5} $88.16\pm0.22$ &\cellcolor{MidnightBlue!5} $80.59\pm0.22$ \\
            &  \texttt{NNM + GeoMed}  &\cellcolor{MidnightBlue!5} \boldmath{$4.89\pm1.21$} & \cellcolor{MidnightBlue!5}$80.28\pm0.15$ &  \cellcolor{MidnightBlue!5}   $58.00\pm38.77$      &\cellcolor{MidnightBlue!5}   $88.27\pm0.15$ &\cellcolor{MidnightBlue!5} $81.02\pm0.25$ \\
            &  \texttt{NNM + Krum} &\cellcolor{MidnightBlue!5} \boldmath{$4.03\pm1.21$} & \cellcolor{MidnightBlue!5} $54.50\pm36.29$ &  \cellcolor{MidnightBlue!5}   $30.58\pm38.76$       &\cellcolor{MidnightBlue!5}   $88.21\pm0.15$  &\cellcolor{MidnightBlue!5} $80.62\pm0.22$ \\
            & \texttt{NNM + CClip} &\cellcolor{MidnightBlue!5} $22.37\pm23.52$ & \cellcolor{MidnightBlue!5} $82.67\pm0.32$ & \cellcolor{MidnightBlue!5} \boldmath{$1.37\pm0.25$} &\cellcolor{MidnightBlue!5}  $87.93\pm0.04$        &\cellcolor{MidnightBlue!5} $82.90\pm0.23$ \\
            \hhline{~------}  
            &  \algname (ours)  &\cellcolor{MidnightBlue!5} $87.54 \pm 0.16$ & \cellcolor{MidnightBlue!5} $87.69 \pm 0.14$ &\cellcolor{MidnightBlue!5} $87.85\pm0.13$   &\cellcolor{MidnightBlue!5} $88.01\pm0.08$ &\cellcolor{MidnightBlue!5} \boldmath{$79.23\pm0.27$} \\
           
            \hline\hline
    \end{tabular}

    }
\end{table}
\begin{table*}[tb]
\renewcommand{\arraystretch}{1.2}
\caption[Final accuracies for CIFAR-10 dataset, $N=20$ ]{Final accuracies for CIFAR-10 dataset, $N=20$, for $f=\{2,4,6\}$, $T=800$. All experiments are run with three random seeds. }
\label{tab:table1}
    \centering
    \resizebox{\textwidth}{!}{
    \begin{tabular}
    {||p{1.2cm}||c||p{2cm}|p{2cm}||p{2cm}| p{2cm}|| p{2cm}| p{2cm}||}
        \hline
        Momentum, Algorithm & Defense & \multicolumn{6}{c||}{Non-IID Data (Dirichlet $\alpha = 0.1$)}\\
        \hline
        \hline
        \multirow{9}{*}{\rotatebox{90}{\parbox{3cm}{\centering$\beta=0$, \vspace{0.3cm} \\ \texttt{FedAVG} ($E=10$)}}}
            & & \multicolumn{2}{c||}{\textbf{$f=2$}} & \multicolumn{2}{c||}{\textbf{$f=4$}} & \multicolumn{2}{c||}{\textbf{$f=6$}} \\ 
            \hhline{~-------} 
             & & \textbf{ALIE} & \textbf{SF} & \textbf{ALIE} & \textbf{SF} & \textbf{ALIE} & \textbf{SF} \\ 
            \hhline{~~------} 
            & \texttt{NNM + Median} &\cellcolor{Salmon!5} \boldmath{$51.31 \pm 2.86 $} &\cellcolor{Salmon!5}  $71.58 \pm 0.79$  & \cellcolor{Maroon!5} \boldmath{$ 32.00\pm1.25 $} &\cellcolor{Maroon!5} $  54.07 \pm 16.78$ &\cellcolor{Brown!5}\boldmath{$17.45 \pm 3.95$} &\cellcolor{Brown!5}{$47.39 \pm 12.38$} \\
            & \texttt{NNM + TrimmedMean} &\cellcolor{Salmon!5} \boldmath{$ 52.75\pm3.28$} &\cellcolor{Salmon!5}  $ 71.20\pm 0.79$  & \cellcolor{Maroon!5} \boldmath{$ 31.68\pm 1.73$}
            &\cellcolor{Maroon!5} $53.91 \pm 15.14$
            &\cellcolor{Brown!5} \boldmath{$15.64 \pm 4.22$} &\cellcolor{Brown!5} {$43.89\pm 15.55$} \\
            & \texttt{NNM + GeoMed} &\cellcolor{Salmon!5} \boldmath{$52.62 \pm 2.71 $}  &\cellcolor{Salmon!5} $71.40 \pm 0.85$  & \cellcolor{Maroon!5} \boldmath{$ 31.89\pm 1.06$}
            &\cellcolor{Maroon!5} $56.89 \pm 13.64$ &\cellcolor{Brown!5} \boldmath{$19.26 \pm 2.40$} &\cellcolor{Brown!5}{$47.51 \pm 11.67$}\\
            & \texttt{NNM + Krum} &\cellcolor{Salmon!5} \boldmath{$50.06\pm3.71$}  &\cellcolor{Salmon!5}  $71.67 \pm 1.00$  & \cellcolor{Maroon!5} \boldmath{$ 29.44\pm 1.47$}
            &\cellcolor{Maroon!5} $ 54.40 \pm 18.92$
            &\cellcolor{Brown!5} \boldmath{$13.84 \pm 1.84$} &\cellcolor{Brown!5}{$ 46.00 \pm 11.63$} \\
            & \texttt{NNM + CClip} &\cellcolor{Salmon!5}\boldmath{$ 53.51\pm 3.53 $ } &\cellcolor{Salmon!5}  $  70.69 \pm 0.62$  & \cellcolor{Maroon!5} \boldmath{$ 31.63\pm 0.70$}
            &\cellcolor{Maroon!5} $ 54.17 \pm 12.47$
            &\cellcolor{Brown!5} \boldmath{$ 20.66\pm 1.26$} &\cellcolor{Brown!5}{$ 42.75 \pm 14.42$} \\
            \hhline{~-------}
            & \algname (ours) &\cellcolor{Salmon!5}$69.86 \pm 2.92 $ &\cellcolor{Salmon!5} \boldmath{ $ 60.03\pm 2.20$}  & \cellcolor{Maroon!5} $68.53 \pm 2.80$ &\cellcolor{Maroon!5} \boldmath{$ 46.54 \pm  8.84 $} &\cellcolor{Brown!5} {$61.94 \pm 4.60$}&\cellcolor{Brown!5} \boldmath{$36.86\pm 13.28$} \\
            \hline
\end{tabular}
    }
\end{table*}

\begin{table*}[tb]
\renewcommand{\arraystretch}{1.3}
\caption[Final accuracies for CIFAR-10 dataset, $N=20$ ]{Final accuracies for CIFAR-10 dataset, $\frac{f}{N}=0.2$, for $N=\{10,20,30,50\}$, $T=800$. All experiments are run with three random seeds. }
\label{tab:table2}
    \centering
    \resizebox{\textwidth}{!}{
    \begin{tabular}
    {||p{1.2cm}||c||p{2cm}|p{2cm}||p{2cm}| p{2cm}|| p{2cm}| p{2cm}||p{2cm}| p{2cm}||}
        \hline
        Momentum, Algorithm & Defense & \multicolumn{8}{c||}{Non-IID Data (Dirichlet $\alpha = 0.1$)}\\
        \hline
        \hline
        \multirow{3}{*}{\rotatebox{90}{\parbox{2cm}{\centering$\beta=0.9$, \vspace{0.3cm} \\  \texttt{FedAVG} ($E=10$)}}}
            & & \multicolumn{2}{c||}{\textbf{$N=10$}} & \multicolumn{2}{c||}{\textbf{$N=20$}} & \multicolumn{2}{c||}{\textbf{$N=30$}} & \multicolumn{2}{c||}{\textbf{$N=50$}} \\ 
            \cline{3-10} 
             & & \textbf{ALIE} & \textbf{SF} & \textbf{ALIE} & \textbf{SF} & \textbf{ALIE} & \textbf{SF} & \textbf{ALIE} & \textbf{SF} \\ 
            \hhline{~~--------} 
            & \texttt{NNM + TrimmedMean} &\cellcolor{Salmon!5}\boldmath{$42.40\pm 5.66$} &\cellcolor{Salmon!5} $ 61.73\pm 4.57 $ & \cellcolor{Maroon!5} \boldmath{$35.35\pm 2.42$}
            &\cellcolor{Maroon!5} $59.02\pm 2.90$
            &\cellcolor{Brown!5} \boldmath{$46.12\pm 4.07$} &\cellcolor{Brown!5} $66.41\pm 3.17$ & \cellcolor{Brown!8} \boldmath{$46.04\pm 1.26$} & \cellcolor{Brown!8}$67.18\pm1.67$ \\
            & \texttt{NNM + CClip} &\cellcolor{Salmon!5}\boldmath{$45.38\pm 4.73$}  &\cellcolor{Salmon!5}  $64.03\pm 3.77 $ & \cellcolor{Maroon!5} \boldmath{$37.05\pm 0.62$}
            &\cellcolor{Maroon!5} $62.86\pm 2.39$
            &\cellcolor{Brown!5} \boldmath{$47.28\pm 4.77$} &\cellcolor{Brown!5} $69.73\pm 2.60$ & \cellcolor{Brown!8} \boldmath{$46.53\pm 1.98$} &\cellcolor{Brown!8} $68.41\pm 3.17$\\
            \hhline{~---------}
            & \algname (ours) &\cellcolor{Salmon!5}$65.60\pm 1.73$  &\cellcolor{Salmon!5}  \boldmath{$62.74\pm 2.95$}  & \cellcolor{Maroon!5} $67.22\pm 3.19$ &\cellcolor{Maroon!5} \boldmath{$54.71\pm 6.09$} &\cellcolor{Brown!5} $71.27\pm 2.89$ &\cellcolor{Brown!5} \boldmath{$63.65\pm 4.74$} & \cellcolor{Brown!8} $70.37\pm 3.62$& \cellcolor{Brown!8}\boldmath{$65.49\pm 2.52$}\\
            
            \hline
            
    \end{tabular}
    }
\end{table*}

\paragraph{Defenses for Evaluation}
For the proposed method \algname, Algorithm \ref{prodigy_algorithm} is used. Denoting by $g^k_i$ the $i$-th component of the gradient vector $\boldsymbol{g}^k$, other robust aggregation rules perform the following operations to output $\hat{\boldsymbol{g}}$:
\begin{itemize}
    \item No Defense:
    $ \texttt{Avg}(\boldsymbol{g}^{1},\dots,\boldsymbol{g}^{N})= \frac{1}{N}\sum_{k=1}^{N}   \boldsymbol{g}^k .$
    \item Coordinate-wise Median~\cite{mediantrimmean}: outputs $\hat{\boldsymbol{g}}$ such that for each entry $\hat{g}_i$,
    \begin{align*}
          \hat{g}_i &= \texttt{Median}(g^1_i,\dots, g^N_i).
    \end{align*}
    \item Coordinate-wise Trimmed Mean~\cite{mediantrimmean}: let $\mathcal{T}_{N-2q}(i)$ denote the multiset obtained from $g^1_i,\dots, g^{N}_i$ by removing the largest and smallest $q$ elements, then
    \begin{align*}
        \texttt{TrimmedMean}_q(\boldsymbol{g}^{1},\dots,\boldsymbol{g}^{N}) = \hat{\boldsymbol{g}} \quad &\text{such that}\\
        \ \hat{g}_i = 
        \frac{1}{N-2q} \sum_{g \in \mathcal{T}_{N-2q}(i)}g.
    \end{align*}
    We choose $q=f$ in the sequel.

    \item Geometric Median~\cite{rfa}: RFA algorithm, with parameters $\nu=0.1$ and $R=3$ rounds for smoothed Weiszfeld algorithm~\cite[Algorithm 2]{rfa} is used,
       $$ \texttt{GeoMed}(\boldsymbol{g}^{1},\dots,\boldsymbol{g}^{N}) = \underset{\boldsymbol{g} \in \mathbb{R}^d}{\arg \min} \sum_{k=1}^{N} \left\|  \boldsymbol{g}-\boldsymbol{g}^k \right\|.$$

    \item Krum~\cite{krum}: For $\mathcal{N}_{N-f-1}(k)$ as defined in \cref{sec:prodigy}:
         $$ \texttt{Krum}(\boldsymbol{g}^{1},\dots,\boldsymbol{g}^{N})= \underset{\boldsymbol{g}^k, k\in [N]}{\arg \min} \sum_{j \in \mathcal{N}_{N-f-1}(k)} \left\|  \boldsymbol{g}^k-\boldsymbol{g}^j \right\|^2.$$

     \item Centered Clipping~\cite{centeredclipping}: for parameters $\tau_l=10$, $L=3$, the algorithm chooses
     \begin{align*}
          &\texttt{CClip}(\boldsymbol{g}^{1},\dots,\boldsymbol{g}^{N}) = \boldsymbol{g}_{L}, \text{ where }\\
         \boldsymbol{g}_{l+1} &= \boldsymbol{g}_{l} + \frac{1}{N} \sum_{k=1}^N(\boldsymbol{g}^{k}-\boldsymbol{g}_{l}) \min \Big(1, \frac{\tau_l}{\left\|(\boldsymbol{g}^{k}-\boldsymbol{g}_{l})\right\|} \Big),
     \end{align*}
     and $\boldsymbol{g}_0$ is initialized by the previous round's aggregation.
\end{itemize}
For a strong comparison, we strengthen the robust aggregation methods we compare against by combining them with the \texttt{NNM} pre-aggregation scheme~\cite{fixbymix}. \texttt{NNM} replaces each gradient with its mix in the following way:
\begin{align*}
     \bar{\boldsymbol{g}}^1, \dots, \bar{\boldsymbol{g}}^{N} &= \texttt{NNM}(\boldsymbol{g}^{1},\dots,\boldsymbol{g}^{N}),\\
     \text{ where} \quad \bar{\boldsymbol{g}}^k &= \frac{1}{N-f} \sum_{i \in \mathcal{N}_{N-f}(k)} \boldsymbol{g}^i,
    \label{eq:nnm}
\end{align*}
with $\mathcal{N}_{N-f}(k)$ as defined in \cref{sec:prodigy}.

\subsection{Experimental Results}

We summarize our experimental results in \cref{tab:bigtable,tab:femnistt,tab:femnistt100,tab:table1,tab:table2}. We first evaluate the accuracy performance of \algname on all considered attacks and compare it to all considered defenses for the set of parameters $N=10$ and $f=3$ for CIFAR-10 and $N=20$ and $f=6$ for FEMNIST \cref{tab:bigtable,tab:femnistt}. To understand the scalability of \algname with respect to $N$ we run experiments with $N=100$ for FEMNIST \cref{tab:femnistt100}. Furthermore, the effect of $N$ and the fraction of Byzantine clients $f$ on defense performance are tested by fixing all parameters and varying $N$ and $f$, respectively \cref{tab:table1,tab:table2}. This ablation study confirms the consistency and validity of the former results.

\paragraph{Impact of Data Heterogeneity} \cref{tab:bigtable} shows the overall comparison of the robust aggregation schemes using IID and non-IID CIFAR-10 datasets, respectively, with $N=10$, and $f=3$, i.e., $30\%$ malicious clients. Under IID data distribution, taking multiple local iterations using \texttt{FedAVG} appears to contribute to the robustness of the schemes as well as to the overall accuracy performance when there is no attack. Similarly, incorporating history through local momentum yields comparable robustness improvements. Therefore, the IID data distribution results reported in \cref{tab:bigtable} indicate that the combination of local momentum and \texttt{FedAVG} with \texttt{NNM} significantly enhances the robustness of the aggregation schemes. Under this setting, the worst-case accuracies of \texttt{Median}, \texttt{TrimmedMean}, \texttt{GeoMed}, \texttt{Krum}, and \algname reach approximately $80\%$, demonstrating that these methods successfully mitigate the attacks.

However, when the data distribution is non-IID, neither \texttt{FedAVG} nor local momentum consistently improves the robustness of these schemes. This is primarily due to the significant divergence among local data distributions, which leads to a large dissimilarity in honest client gradients. This dissimilarity gap is further exacerbated, enabling an attacker to manipulate many of the defenses by colluding, such as in ALIE and FOE attacks. While a degradation in accuracy is natural in such a challenging non-IID setting, it can be observed that \algname is significantly more robust than other schemes. Eventually, all other evaluated robust aggregation schemes exhibit lower worst-case accuracy than \algname, frequently performing no better than random guessing (approximately $10\%$).

To further investigate non-IID data distribution, the experiments are conducted on the FEMNIST dataset, where the source of heterogeneity is feature skew, and local dataset sizes are considerably more balanced than the strongly non-IID CIFAR-10 datasets. \cref{tab:femnistt,tab:femnistt100} show that \algname consistently maintains robustness against all attack scenarios while other defenses yield substantially lower worst-case accuracies due to failure. The large standard deviation of the performances indicates inconsistent and unreliable robustness, which is mainly attributable to the limited number of repetitions with random seeds.

\paragraph{Impact of Fraction of Byzantine Clients}

To assess the impact of varying fractions of malicious clients, we fix the total number of clients at $N=20$ and evaluate performance under $10\%$, $20\%$, $30\%$ malicious participation on CIFAR-10, as reported in \cref{tab:table1}. For the comparison, we focus only on ALIE and SF attacks since ALIE performs the most successful attack, frequently circumventing all defenses except \algname, and SF is the worst case attack against \algname based on \cref{tab:bigtable,tab:femnistt,tab:femnistt100}. While the absolute attack power increases with a higher percentage of malicious clients, \algname consistently maintains the highest worst-case accuracy. 

\paragraph{Impact of Number of Clients}

We evaluate the impact of the total client number on attack and defense performances while keeping the malicious fraction fixed. For this evaluation, \algname is compared with \texttt{TrimmedMean}, which we choose randomly as a representative of other defenses due to similar robustness performances of those defenses. Additionally, we choose \texttt{CClip} due to its distinct approach and different robustness performance. \cref{tab:table2} presents results from experiments with varying total numbers of clients. 
Model performance remains generally consistent across different numbers of participating clients, except for some fluctuations. These fluctuations primarily stem from the diverse and highly non-IID data distributions, which vary substantially with the number of clients.

\begin{figure}[!htb]
     \centering
     \includegraphics[width=0.75\columnwidth]{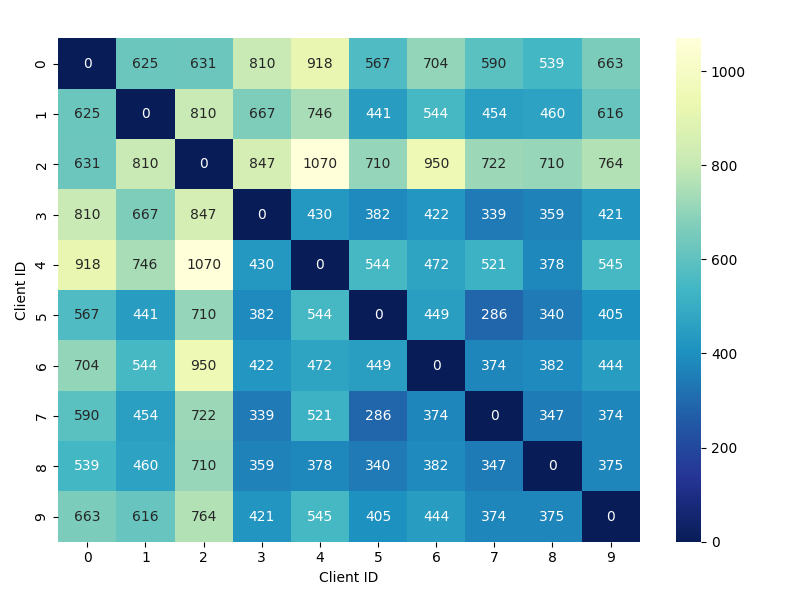}

     \caption{Pairwise squared distances between gradients, based on the CNN model training on CIFAR-10 data, $N=10$, $f=3$ in communication round $t=100$. Non-IID distribution with $Dir_N(0.1)$ is considered. The gradients are collected from the experiment with SF attack and with \algname defense, using \texttt{FedAVG}. Byzantine client IDs are $\{0,1,2\}$.}
    \label{fig:similaritymetriccomp}
\end{figure}

\paragraph{Limitations under Heterogeneity}
Interestingly, we observe that SF attack is the most challenging attack for \algname, leading to the lowest accuracy we observed throughout our experiments for \algname. To better understand how SF in the case of severe heterogeneity can confuse defense methods, we check the pairwise distance matrix in \cref{fig:similaritymetriccomp}. In the pairwise distance matrix a malicious gradient can sometimes have a smaller distance to an honest gradient than another honest gradient. For example, in \cref{fig:similaritymetriccomp}, Client 4 exhibits a greater distance to Client 5 than Client 1 does, despite Client 1 being malicious. Even when a Byzantine client sends a gradient in the opposite direction of its computed gradient, it may still appear closer to another honest gradient, while honest gradients themselves deviate substantially, which highlights the difficulty of a heterogeneous setting.

We remark that the learning under SF attack experiences fluctuations when local momentum is not applied, for all considered defenses. This is most evident when $N=20$, $f=4$ cases in \cref{tab:table1} and \cref{tab:table2} are compared. When local momentum is used, the learning behavior becomes more stable, resulting in low fluctuation of the final accuracies. Also we observe in our experiments that, under severe data heterogeneity, applying $\texttt{Avg}(\cdot)$ as an aggregation mechanism, referred to as No Defense, proved to be the most resilient defense against the SF attack when local momentum is integrated (see \cref{tab:bigtable,tab:femnistt,tab:femnistt100}). We ascribe this observation to the fact that it consistently aggregates gradients without favoring any particular direction. Given that the majority is honest, averaging makes the aggregation result most faithful to the honest model updates, whereas defenses may occasionally misidentify malicious updates as honest.

\section{Conclusion}

We introduced \algname, a Byzantine-robust aggregation method that employs a dual scoring mechanism based on both gradient proximity and dissimilarity. Through extensive experimentation across diverse settings, we show that \algname consistently outperforms existing defense mechanisms in terms of worst-case accuracy. Even in challenging non-IID settings, where many prominent defenses fail, it maintains an acceptable accuracy. The main novelty of \algname lies in its ability to mitigate the colluding power of adversaries by penalizing over-proportional gradient similarity. The exploration of optimal attack strategies by colluding adversaries, that maximize the disruption of the learning process while evading detection, is the subject of future research. The best defense performance is achieved when the number of Byzantine clients, $f$, is accurately estimated. Accordingly, future work may focus on integrating an estimation strategy for $f$ into \algname. Our future work also includes investigating theoretical guarantees for the robustness of \algname, and further demonstrating its applicability across diverse learning tasks.

\balance
\bibliographystyle{IEEEtran}
\bibliography{references}

\end{document}